\documentclass[conference,letterpaper]{IEEEtran}

\usepackage{graphicx}
\usepackage{amssymb,amsmath}
\usepackage{cite} 
\usepackage{booktabs}
\usepackage[font=footnotesize,labelformat=simple]{subcaption}
\usepackage[font=footnotesize]{caption}

\usepackage{hyperref}
\hypersetup{bookmarksopen,bookmarksnumbered,
pdfpagemode=UseOutlines,
colorlinks=true,
linkcolor=blue,
anchorcolor=blue,
citecolor=blue,
filecolor=blue,
menucolor=blue,
urlcolor=blue
}

\newcommand{\xxnote}[3]{}
\ifx\hidenotes\undefined
  \usepackage{color}
  \renewcommand{\xxnote}[3]{\color{#2}{#1: #3}}
\fi

\begin{document}
{\title{
Improved Proximity, Contact, and Force Sensing via Optimization of Elastomer-Air Interface Geometry\\
}}
\author{
\IEEEauthorblockN{
Patrick E. ~Lancaster,
Joshua R. ~Smith,
Siddhartha S.~Srinivasa
}
\IEEEauthorblockA{
School of Computer Science \& Engineering, University of Washington\\
\{planc509, jrs, siddh\}@cs.uw.edu
}
}

\maketitle

\begin{abstract}

We describe a single fingertip-mounted sensing system for robot manipulation that provides proximity (pre-touch), contact detection (touch), and force sensing (post-touch). The sensor system consists of optical time-of-flight  range measurement modules covered in a clear elastomer. Because the elastomer is clear, the sensor can detect and range nearby objects, as well as measure deformations caused by objects that are in contact with the sensor and thereby estimate
the applied force. We examine how this sensor design can be improved with respect to invariance to object reflectivity, signal-to-noise ratio, and continuous operation when switching between the distance and force measurement regimes. By harnessing time-of-flight technology and optimizing the elastomer-air boundary to control the emitted light's path, we develop a sensor that is able to seamlessly transition between measuring distances of up to 50 mm and contact forces of up to 10 newtons. Furthermore, we provide all hardware design files and software sources, and offer thorough instructions on how to manufacture the sensor from inexpensive, commercially available components.

\end{abstract}

\section{Introduction}

Endowing robots with finger tips that are sensitive to proximity, contact, and force (PCF) provides them with sensor feedback during critical moments of manipulation. Proximity data helps the robot localize an object, while contact and force data allows the robot to adjust how it is influencing an object, such as loosening its grip, pushing forward harder, or detecting slip. PCF sensors avoid the pitfalls of occlusion, unintentional object displacement, and non-zero sensing range that characterize head-mounted sensors, tactile-only sensors, and proximity-only sensors respectively. Recognizing combined PCF sensing as a relatively new concept, we focus on developing methods to further improve its capabilities and thereby its utility in robot manipulation. 

We choose to focus on the optical sensing modality because it produces accurate measurements over a wide range. In particular, Patel et. al. \cite{patel2017integrated} describe an innovative design for an optical PCF sensor. They place a clear elastomer over an optical proximity module. The ranging module can measure the distance to a nearby object because the elastomer is clear. To infer the force applied by an object in contact with the sensor, the elastomer is modelled as a spring that has been compressed by the measured distance. 

\begin{figure*}[!t]
\centering
\includegraphics[width=\linewidth]{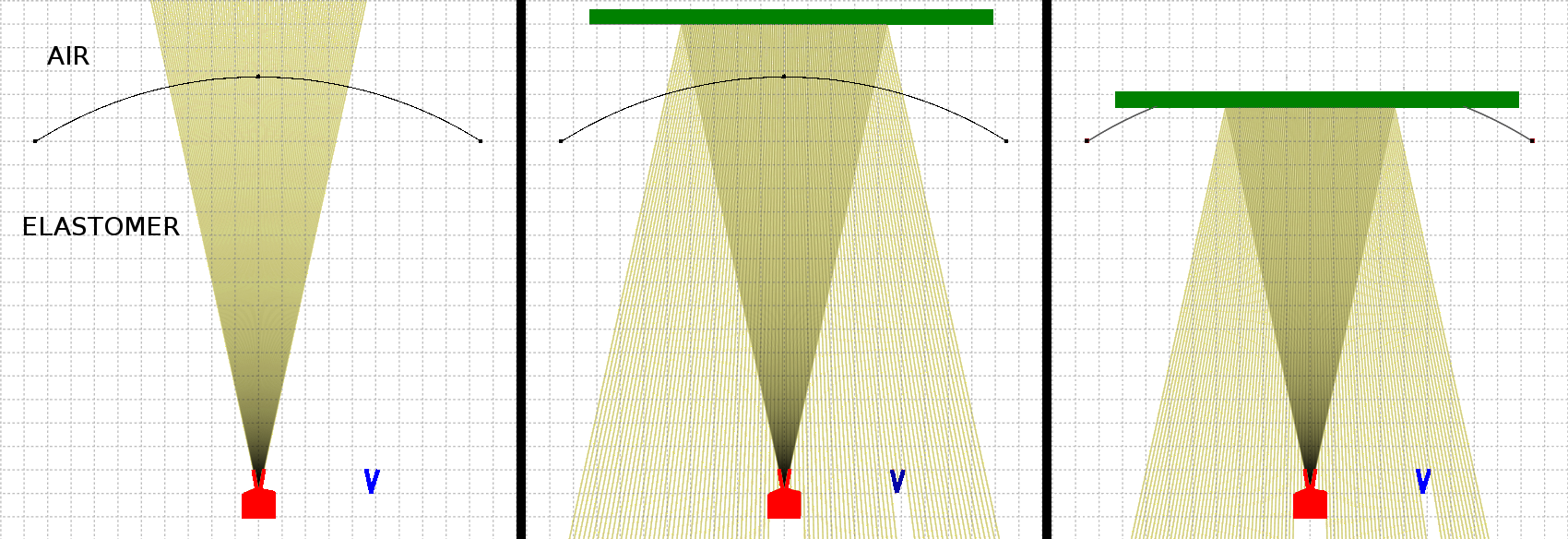}
\caption{The path of infrared light emitted by the sensor when using a rounded elastomer-air boundary. The emitter is shown in red, the receiver in blue, the boundary in black, and a target object in green. All relevant metrics are to scale including the angle of light emission, transmitter and receiver aperture, curvature of the boundary, and spatial relationship between the transmitter, receiver and boundary. Each grid unit is equivalent to one millimeter \cite{tu2018ray}. \\ \textbf{Left:} No object is present. While most of the light transitions through the boundary, some of it reflects off of the boundary and back into the transmitter. \textbf{Middle:} Object is in proximity. \textbf{Right:} Object makes contact and deforms the boundary.}
\label{fig:summary}
\end{figure*}

While \cite{patel2017integrated} considers a number of design parameters, we explore other directions that further optimize the performance of this type of PCF sensor. First, our sensor uses time-of-flight (in addition to intensity) of reflected infrared light to measure distance. This allows our sensor's proximity measurements to be invariant to object surface reflectivity, which is a drawback that \cite{patel2017integrated} suffered from. Another area in which we improve upon \cite{patel2017integrated} is that their implementation operates in two separate modes. When measuring force, an electrical current configuration that optimizes sensitivity is used. However, when measuring distance, the infrared light reflecting from the elastomer-air boundary reaches the receiver and decreases the signal to noise ratio. We hypothesize that the second configuration that uses a greater amount of current is required to achieve an acceptable signal to noise ratio. We avoid having two different configurations by optimizing the elastomer-air boundary to control the path of reflected light (Fig. \ref{fig:summary}). 

\begin{figure}[!t]
\centering
\includegraphics[width=\linewidth]{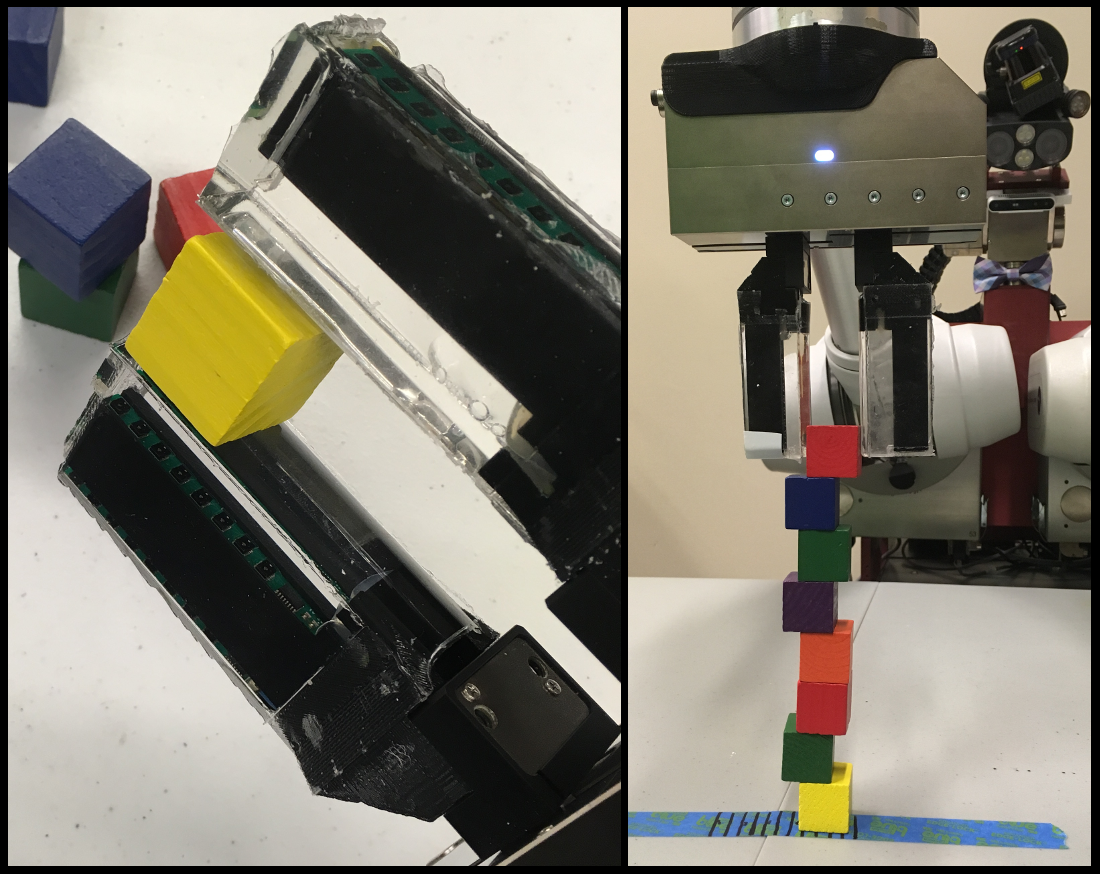}
\caption{\textbf{Left:} Two optical time-of-flight proximity, contact, and force sensors with rounded elastomer-air interfaces that have been fully integrated into a WSG50 Weiss Gripper. \textbf{Right:} The robot uses the sensors to \textbf{unstack} blocks.}
\label{fig:cube_grab}
\end{figure}

In this work, we make the following contributions:
\begin{itemize}
\item Design of an improved PCF sensor (Fig. \ref{fig:cube_grab}) that seamlessly transitions between measuring distances of up to 50 mm and contact forces of up to 10 newtons.
\item Publicly available hardware sources, software sources, and thorough instructions  for fabrication of the sensor from inexpensive, commercially available components - which can all be found at \href{ https://bitbucket.org/opticalpcf/}{https://bitbucket.org/opticalpcf/}.
\item Demonstration of how the sensor can help the robot perform the delicate task of unstacking blocks (Fig. \ref{fig:cube_grab}).
\end{itemize}

\section{Related Work}

Grasping, a seemingly simple manipulation task, still presents a challenge for robots because of the difficulty involved in perceiving the pose of the object of interest with respect to the robot's end effector. Often, this localization is achieved through computer vision techniques that receive data from a head mounted camera or depth sensor \cite{azad2007stereo}\cite{saxena2008robotic}\cite{stuckler2013efficient}\cite{collet2009object}\cite{katz2008manipulating}.  However, objects of interest can become occluded from view by the hand. In this section, we focus on reviewing how past works augment the robot's perception system with finger mounted sensors that are robust to occlusion, generate low-dimensional data, and provide more accurate measurements due to their closer proximity to the target.  

A number of different technologies have been used to achieve tactile sensing \cite{yousef2011tactile}\cite{hughes2015texture}. Zhang et. al. \cite{zhang2013soft} measure forces through the deformation of optical fibers embedded in an elastomer. Other works achieve high resolution contact localization by considering how the path of emitted light is altered by deformations of the sensor's elastomer \cite{li2014localization} \cite{piacenza2017accurate}. By placing a clear elastomer on top of a camera, Yamaguchi and Atkeson \cite{yamaguchi2016combining} are able to estimate force, segment objects, and detect slip. The utility of these types of sensors is demonstrated in a bayesian framework developed by Petrovskaya and Khatib \cite{petrovskaya2011global}, in which they use contact data to increasingly refine the pose estimate of an object. One drawback of contact based sensing is that it requires touch to detect an object, potentially resulting in incidental displacement.

Proximity sensors operate at a range intermediate to that of head-mounted sensors and tactile sensors. Balek and Kelley \cite{balek1985using} provide one of the earliest studies on how optical sensors can be integrated into robot grippers for obstacle avoidance, localization, and object recognition. More recent studies of optical based sensors have explored their application to grasping \cite{hsiao2009reactive}\cite{guo2015transmissive}\cite{jiang2015fiber}, sensor fusion \cite{cox2017merging}\cite{konstantinova2016fingertip}\cite{koyama2013pre}, slip detection \cite{maldonado2012improving}, and sequential manipulation \cite{yang2017pre}. Compared to other sensing modalities, optical sensors typically provide more accurate measurements over a wider range. However, they may fail to detect objects that are transparent or highly specular. In contrast, electric field sensing is relatively invariant to object surface properties, but is only sensitive to materials that have a dielectric constant significantly different from that of air. Electric field sensors \cite{wistort2008electric}\cite{smith2007electric}\cite{mayton2010electric}\cite{muhlbacher2015responsive} transmit an alternating current from one or more electrodes, and then measure how objects affect the displacement current flowing into one or more receive electrodes. While electric field and optical sensing methods actively emit a signal and measure how it is distorted or reflected, acoustic proximity sensing is a passive technique that has not been extensively explored. The sensor consists of a cylindrical tube with a microphone at one end. The microphone measures the resonant frequency of the tube, which is modulated as objects approach it \cite{jiang2012seashell}\cite{huang2015sensor}.

Koyama et.al \cite{koyama2018high} develop a pre-cursor to optical PCF sensors. Specifically, they report a high speed, high accuracy sensor that measures both proximity and contact, but not force. In contrast, \cite{patel2017integrated} reduces the physical footprint of the sensor by using a ranging module that is packaged as an integrated chip. This facilitates the creation of sensor arrays and the ability to measure force, albeit at the cost of accuracy and speed. In later work, they demonstrate how this sensor can be combined with vision to achieve more robust grasping \cite{patel2017improving}.  Our work is more similar to \cite{patel2017integrated} in that we demonstrate the feasibility of using a commercially available ranging module to measure proximity, contact, and force. Our sensor uses time of flight technology to improve its robustness to varying object reflectivity. Also, unlike \cite{patel2017integrated}, our sensor does not need to be reconfigured when switching between the distance and force measurement regimes. By optimizing the elastomer-air boundary, we can control the path of internally reflected light, resulting in improved signal-to-noise ratio.





\section{Improved Proximity and Force Sensing}

Conventional reflective optical proximity sensors are severely affected by the surface properties of target objects because they directly measure the intensity of received light. Furthermore, while installing a clear elastomer over top of a proximity sensor establishes a mechanism for measuring force, infrared light that reflects from the elastomer-air boundary into the receiver reduces the signal-to-noise ratio. We employ a ranging module that can measure the time-of-flight of infrared light to remove the effect of surface reflectivity. We also design the geometry of the elastomer-air boundary in order to control the path of internally reflected light. In this section, we first characterize both the time-of-flight and the intensity outputs of the ranging module. Then we detail the theory and practice of how the elastomer-air boundary can be optimized to guide internally reflected light away from the receiver.

\subsection{Time-of-Flight Sensor Operation}
\begin{figure}[b]
\centering
\includegraphics[width=0.45\textwidth]{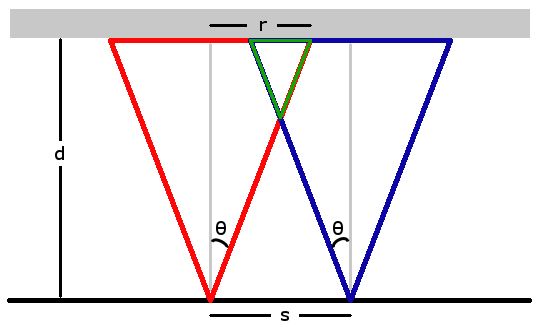}
\caption{ A side view of the sensor. The emitter produces a cone of photons as illustrated in red. The receiver cone is illustrated in blue. Their intersection forms the green viewing cone of the sensor, which defines the portion of photons that reflect off of the gray target object and into the receiver
}
\label{fig:emitter_receiver}
\end{figure}

Our sensor uses the STMicroelectronics VL6180X Proximity Sensing Module to measure the distance travelled and intensity of emitted infrared light. Fig. \ref{fig:emitter_receiver} illustrates the cone of emitted photons and the portion of them that are reflected from the object at distance $d$  and back into the receiver. The half field-of-view angle $\theta$ is the same value for both the emission and reception cones, causing them to have the same radius $r$. The emitter and receiver are separated by a distance $s$. 

\begin{figure}[!t]
\centering
\includegraphics[width=0.45\textwidth]{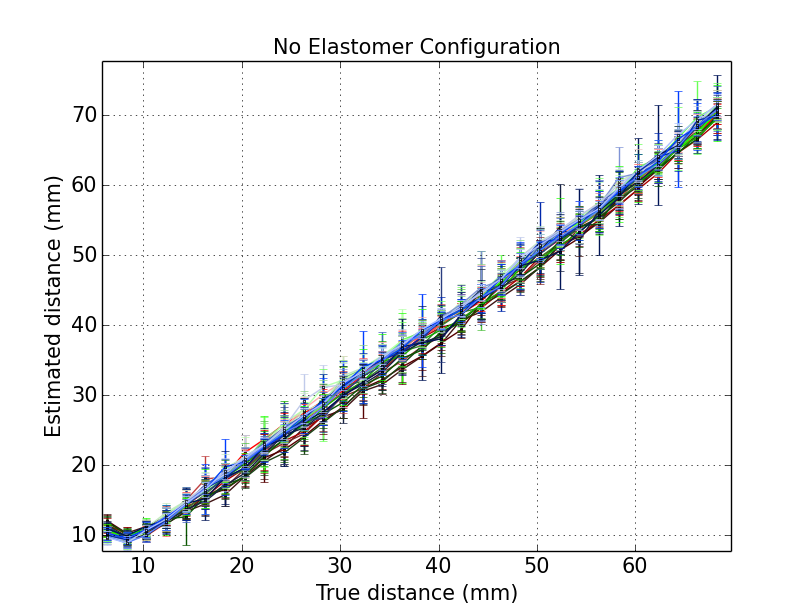}
\caption{ The range output of the VL6180X. The color of each series corresponds to the color of the target.
}
\label{fig:bare_range_color_dist}
\end{figure}

Distance is estimated by measuring the average time that it takes for emitted photons to reflect off of a target object and then return to the receiver. Fig. \ref{fig:bare_range_color_dist} shows distance measurements for different colored targets - dark, light and in-between shades of red, green, and blue. The measurements are fairly linear with respect to the true distance, and typically have an error on the order of a few millimeters.

The sensor simultaneously provides an estimate of the intensity of the reflected infrared light by measuring the rate at which photons are received. This phenomenon can be modelled by considering the geometry of the interaction between the sensor and target object. Intensity will be directly proportional to the ratio of the number of photons received to the number of photons emitted. This ratio is proportional to the ratio of the area of the object within the viewing cone $A_{view}$ (i.e. the intersection of the emitter and receiver cones) to the area of the object within the emitter cone $A_{emit}$:
\begin{center}
$\hat{I} \propto \frac{A_{view}}{A_{emit}}$
\end{center}
\begin{center}
$A_{view} = \frac{s}{2}\sqrt[]{4r^2-s^2}$
\end{center}
Given that $A_{emit} = \pi r^2$ and $r = d\tan(\theta)$, we obtain:
\begin{center}
$\hat{I} \propto \frac{s}{2} \frac{\sqrt[]{4 d^2 \tan^2\theta-s^2}}{\pi d^2 \tan^2\theta}$ 
\end{center}

By combining constants and adding an offset term, we obtain the following expression:

\begin{center}
$\hat{I} = \kappa \frac{ \sqrt[]{d^2 -\zeta^2}}{d^2}+\chi$ 
\end{center}

\begin{figure}[!hb]
\centering
\includegraphics[width=0.45\textwidth]{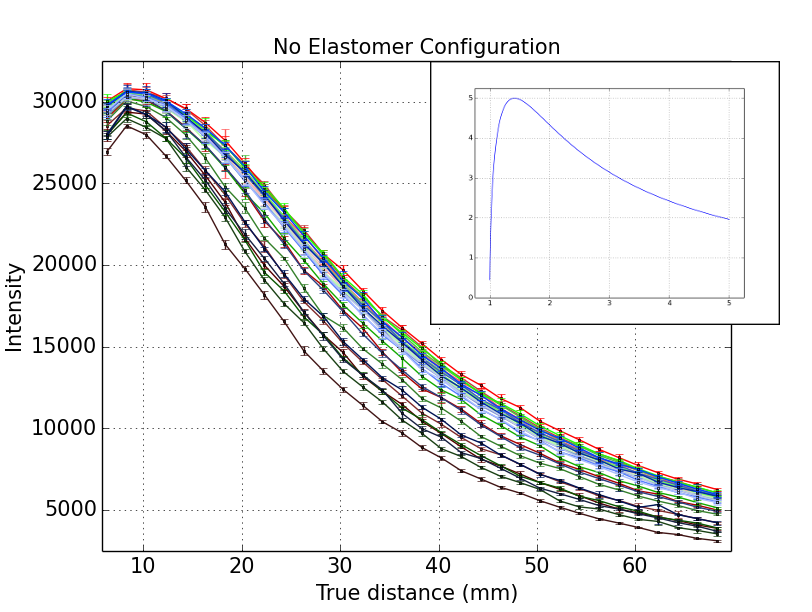}
\caption{ The intensity output of the VL6180X. The color of each series corresponds to the color of the target. Inset shows derived functional form.
}
\label{fig:bare_intensity_color_dist}
\end{figure}

Using the same targets previously mentioned, intensity measurements are shown in Fig. \ref{fig:bare_intensity_color_dist}. We observe that the data does approximate the derived functional form above.

\begin{figure*}[!t]
\centering
\includegraphics[width=\linewidth]{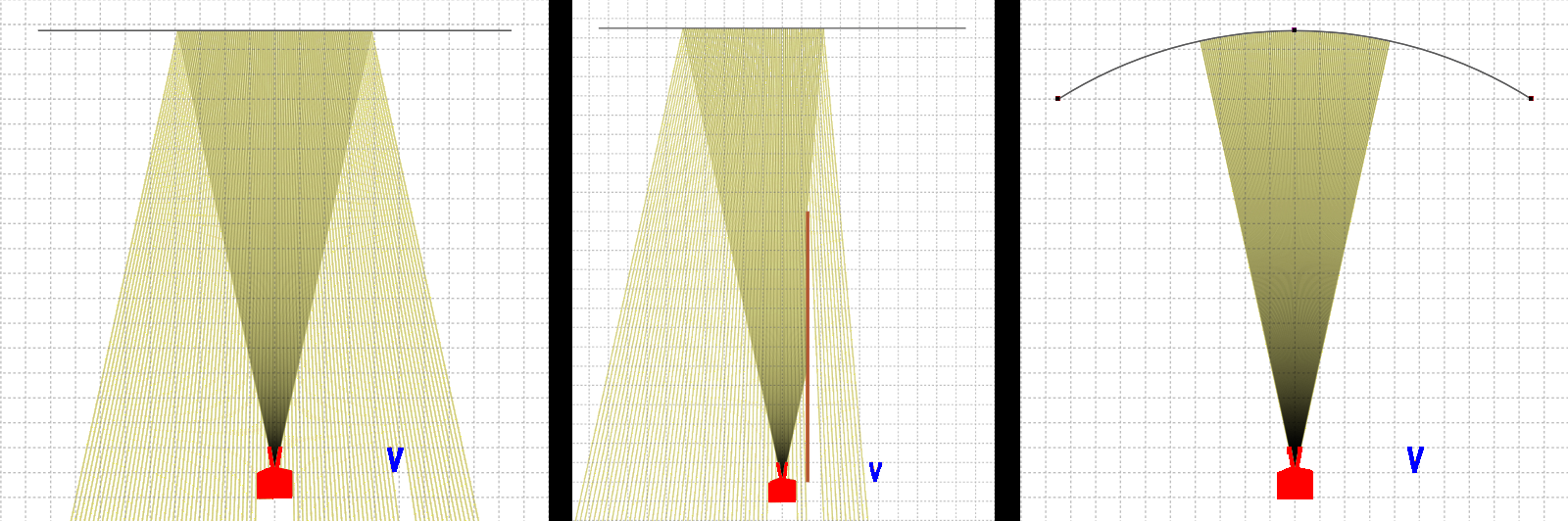}
\caption{\textbf{Left:} Flat configuration. \textbf{Middle:} Blocker configuration \textbf{Right:} Circular arc configuration. The path of emitted light rays for each of the examined elastomer-air boundary configurations. Note that only reflected rays are shown, refracted rays are omitted for clarity. The emitter is shown in red, the receiver in blue, the boundary in black, and the blocker in brown. All relevant metrics are to scale including the angle of light emission, transmitter and receiver aperture, curvature of the boundary, and spatial relationship between the transmitter, receiver, and boundary. Each grid unit is equivalent to one millimeter \cite{tu2018ray}. }
\label{fig:light_rays}
\end{figure*}

\begin{figure}[!hb]
\centering
\includegraphics[width=0.45\textwidth]{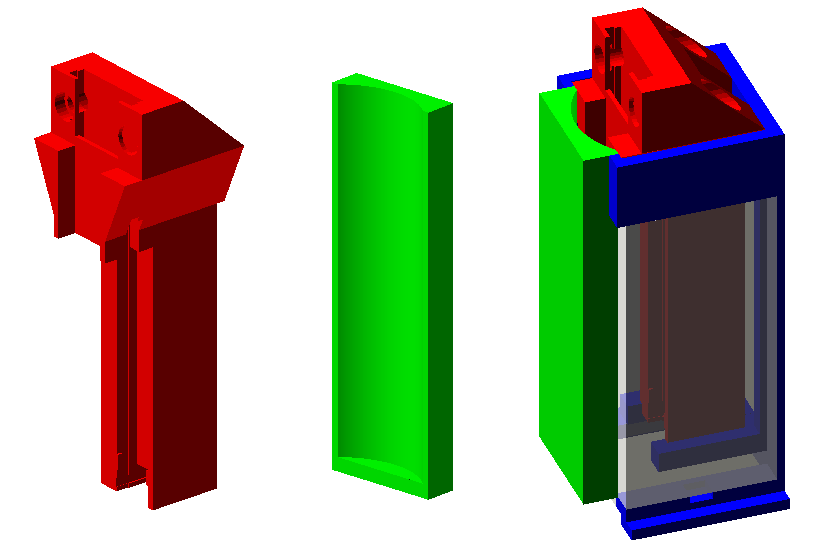}
\caption{ \textbf{Left:} Case for mounting sensor to Weiss gripper. \textbf{Middle:} Portion of mold that forms the circular arc elastomer-air boundary. \textbf{Right:} The assembled mold. Transparent planes correspond to pieces of acrylic.}
\label{fig:mold_parts}
\end{figure}

\subsection{Elastomer-Air Boundary Optimization}
If not treated carefully, light reflecting from the elastomer-air boundary will become a source of interference. Fig. \ref{fig:light_rays} illustrates this effect when a naiive flat geometry is used. One solution to prevent infrared light from reaching the receiver is to simply make the distance between the sensor and the boundary smaller. However, doing so would lead to insensitivity in the force regime because no light would be able to reach the receiver. Fig. \ref{fig:light_rays} presents two other possible solutions. First, one could place a physical blocker between the emitter and receiver. Alternatively, the geometry of the boundary could be shaped into a circular arc to reflect light back into the transmitter, and therefore away from the receiver. In order to best focus the reflected light into the transmitter, the transmitter should be placed at the focus of the circular arc, i.e. one radius away from the boundary.  This distance $d$ defines the thickness of the elastomer. The remainder of this section will detail the fabrication of the circular arc configuration.

Our sensor is fabricated by first 3D printing a case that can be mounted to the WSG-50 Weiss gripper. We place the case and electronics into a mold as shown in Fig. \ref{fig:mold_parts}. Note that we print the mold such that its main axis is vertical so that the circular geometry is not limited by the layer height. The thickness $d$ of the elastomer should be chosen to correspond to the operation region in which the ranging module is most sensitive. We found that a thickness of 17.75 millimeters works well for both the flat and circular arc geometries. We increase the thickness for the blocker configuration to approximately 23.5 milimeters so that light can still reach the receiver when the boundary is deformed. The elastomer is formed by pouring polydimethylsiloxane (PDMS) into the top of the mold. We use a mixing ratio of 30:1 to achieve a low Young's modulus, i.e. an elastomer that is more easily compressed by applied force. 

Achieving an elastomer-air boundary that is optically clear is paramount to maintaining a high signal-to-noise ratio in the proximity ranging regime. This requires that the part of the mold that determines the boundary be smooth. The flat configuration, blocker configuration, and the implementation from \cite{patel2017integrated} use a flat piece of acrylic for this part of the mold to achieve an optically clear boundary. However, when the elastomer-air boundary requires a curved geometry, obtaining optical clarity in an inexpensive way is more difficult. Simply 3D printing this section of the mold will produce unsatisfactory results because its layer-wise nature will cause unsmoothness. This is demonstrated in \cite{patel2018integrated}, in which the work of \cite{patel2017integrated} is adapted for a Kinova gripper. The resulting elastomer-air boundary is opaque compared to the original implementation.

To manufacture an optically clear rounded elastomer-air boundary, we apply a number of post-processing steps after 3D printing the mold. We apply Smooth-On XTC3D Epoxy to this portion of the mold. Specifically, we apply two coats of the epoxy to the mold, sand it with 400 grit sandpaper, and then apply one more coat of epoxy. The resulting surface is sufficiently smooth to produce an optically clear boundary.


\begin{figure*}[!t]
\centering
\includegraphics[width=\linewidth]{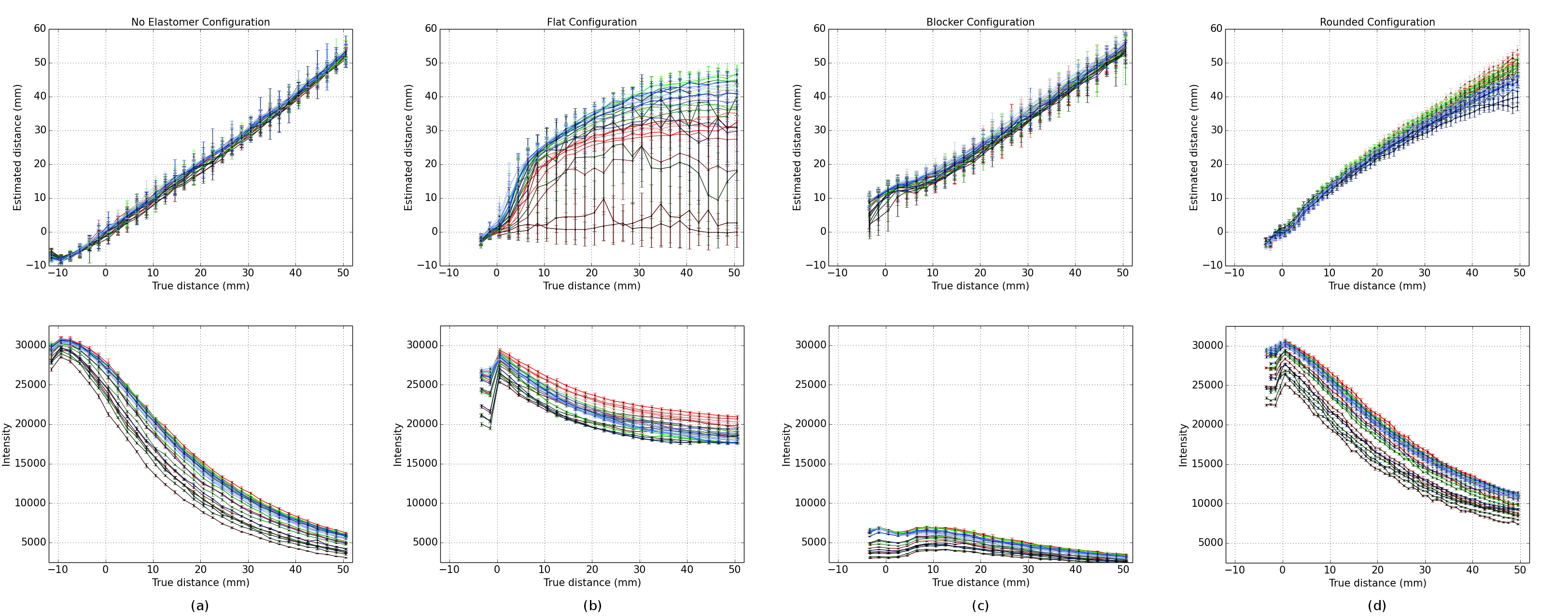}
\caption{\textbf{Far Left:} Output of VL6180X sensor (no elastomer). \textbf{Middle Left:} Output of flat configuration. \textbf{Middle Right:} Output of blocker configuration. \textbf{Far Right:} Output of rounded configuration. Range (upper row) and intensity (lower row) outputs for each configuration in the proximity regime. The zero point corresponds to the point at which an object makes contact with the sensor. The color of each series corresponds to the color of the target. }
\label{fig:color_dist}
\end{figure*}


\section{Experiments}
\begin{figure}[!t]
\centering
\includegraphics[width=0.45\textwidth]{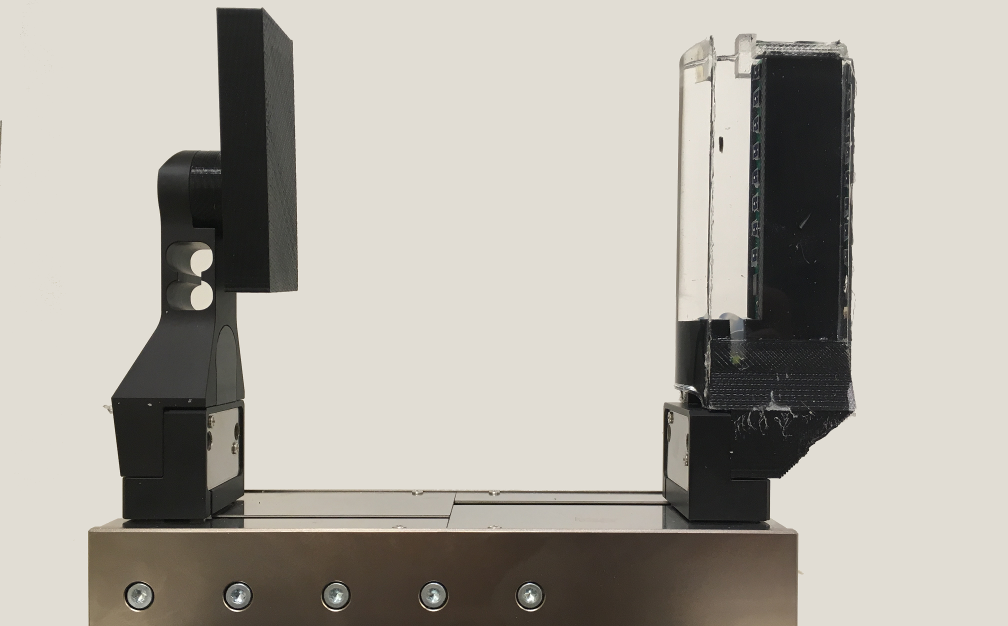}
\caption{ The Weiss Force Measurement Finger attached to the left finger of the Weiss hand and the sensor with rounded configuration attached to the right finger. The sensor's outputs are measured as the gripper width is varied.
}
\label{fig:experiment_setup}
\end{figure}

We compare the performance of the flat, blocker, and circular-arc configurations in both the proximity and force measurement regimes. As shown in Fig. \ref{fig:experiment_setup}, the experimental setup consists of the sensor mounted to one finger of the Weiss hand, while the Weiss Force Measurement Finger (FMF) is mounted to the other. Different colored targets are attached to the plane of the FMF finger in order to measure the sensor's sensitivity to surface reflectance. The targets include dark, light (17\%  and 85\% lightness in the HLS color space respectively), and in between shades of red, green, and blue. Ground truth distance is provided by the hand's motor encoder, which has a resolution of 0.1 mm. The FMF finger provides ground truth forces with plus or minus three percent accuracy.

\begin{figure*}[!t]
\centering
\includegraphics[width=\linewidth]{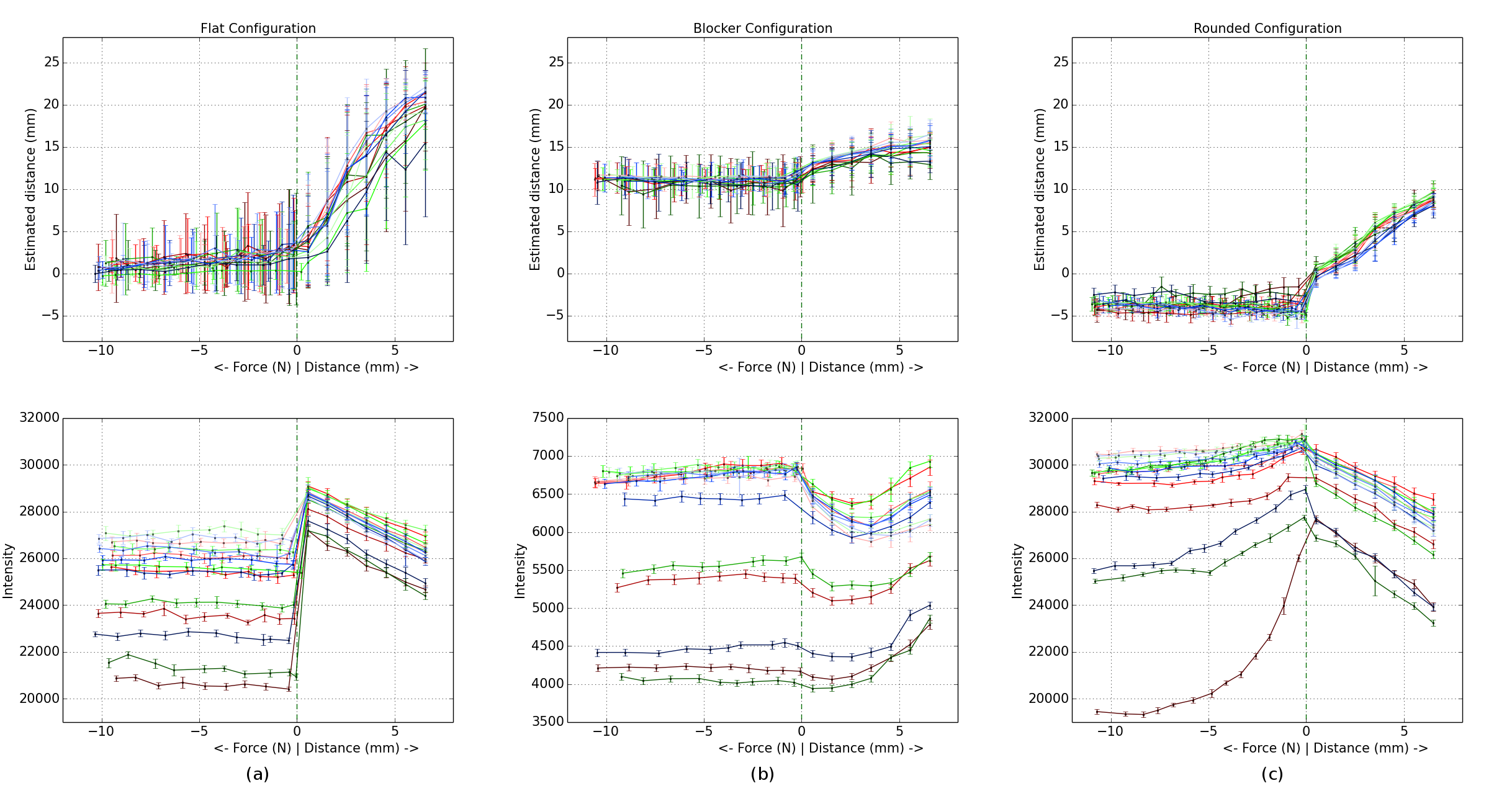}
\caption{\textbf{Left:} Output of flat configuration. \textbf{Middle:} Output of blocker configuration. \textbf{Right:} Output of rounded configuration. Range (upper row) and intensity (lower row) outputs for each configuration in the force regime. The zero point (denoted by a vertical dashed line) corresponds to the point at which an object makes contact with the sensor. The color of each series corresponds to the color of the target.}  
\label{fig:color_force}
\end{figure*}


We first analyze the performance of all three configurations in the proximity regime. Fig \ref{fig:color_dist}(a) illustrates the performance of the sensor when it is not covered by elastomer. This represents the best performance that we could hope to achieve because placing elastomer on top of the sensor will only degrade the emitted signal and cause interference. 
In Fig. \ref{fig:color_dist}(b), we can see that the flat configuration sensor's range output is highly non-linear and influenced by surface reflectivity, and its intensity output has less dynamic range. In Fig. \ref{fig:color_dist}(c), the blocker configuration's range output has accurate, linear behavior throughout most if its range, but becomes non-linear in the region approximately one centimeter before contact. Also, its intensity output is relatively flat. The rounded configuration in Fig. \ref{fig:color_dist}(d) is also linear and accurate up until the four to five centimeter range at which darker colors result in less than linear behavior. Furthermore, its intensity measurements are comparable to that of the elastomerless configuration. The data suggest that of the three configurations, the rounded configuration performs best. In particular, this configuration is preferable to the blocker configuration because it is more likely that accurate range measurements will be required just prior to contact rather than when objects are far away. 

To detect contact, we focus on the range output shown in Fig. \ref{fig:color_force}. In general, all three configurations decrease as the sensor transitions from the pre-contact regime to the post-contact regime. This decrease is most dramatic and the least noisy for the circular-arc configuration. The flat configuration also has the potential to detect contact well based on the intensity data shown in Fig. \ref{fig:color_force}(a). However, it may be limited by the intensity output's sensitivity to target reflectivity.

\begin{figure}[!b]
\centering
\includegraphics[width=0.45\textwidth]{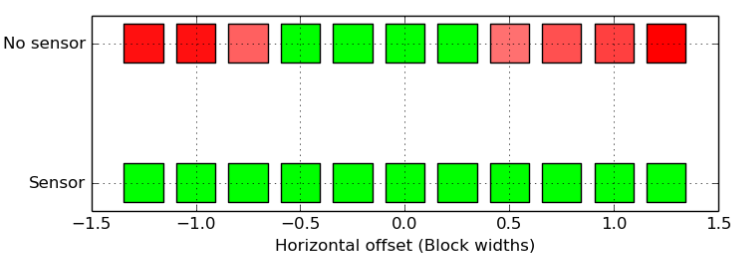}
\caption{Green indicates that the robot successfully unstacked all of the blocks at the corresponding position with the specified method. Red indicates failure, where darker shades indicate that fewer blocks were successfully unstacked. One block width is one inch. Each stack consists of eight blocks.
}
\label{fig:block_offset}
\end{figure}

As shown in Fig. \ref{fig:color_force}, the range output for all of the configurations is relatively flat and/or noisy throughout the force regime. Therefore, force measurements will primarily be derived from the intensity output. The intensity values for the flat (Fig. \ref{fig:color_force}(a)) and blocker (Fig. \ref{fig:color_force}(b)) configurations are both flat. In contrast, the circular arc configuration (Fig. \ref{fig:color_force}(c)) monotonically decreases as more force is applied. We also observe that its sensitivity depends on the reflectivity of the target, where less reflective targets result in greater sensation.

After finding that the sensor with a rounded elastomer-air boundary is the best performing configuration in both the proximity and force regimes, we demonstrate its utility for robots operating in uncertain environments. The robot must deconstruct a stack of eight one-inch cubes that has been offset from its expected location. As shown in Fig. \ref{fig:block_offset}, we examine how the robot performs both with and without using sensor feedback with the stack placed at various horizontal offsets. When not using sensor feedback, the robot's performance degrades as the offset grows, with the most typical failure being one finger pushing a block off of the stack before the other finger can make contact. By using the sensors mounted to its gripper as shown in Fig. \ref{fig:cube_grab}, the robot can servo its gripper to be centered around the block and robustly unstack it.


\section{Conclusion}
Time-of-flight technology and elastomer-air boundary optimization can drastically improve the capabilities of optical PCF sensors. We observe that using time-of-flight information allows the sensor to perform distance measurements that are robust to target reflectivity in the proximity regime. Furthermore, after considering three different possible elastomer configurations, we found that a circular-arc boundary prevents internally reflected light from reaching the receiver when operating in the proximity regime. As a result, the sensor accurately measures distance and detects contact. This configuration also has the best performance in the force regime, but its measurements are affected by target surface reflectivity.

One limitation of the sensor is that some measure of the reflectivity of the target must be known in order to correlate an intensity measurement with an observed force. In future work, we intend to characterize the surface reflectivity of the target in the proximity regime, and then use this information in the force regime to make accurate force measurements. The robot could first measure the object at some distance to obtain a range measurement and an intensity measurement. Based on the data in Fig. \ref{fig:color_dist}, we believe that this pair of data is sufficiently discriminative to characterize the target's surface properties. Using this information, the sensor can lookup the force that corresponds to a measured intensity value once the target has made contact. With this capability, future work will focus on how this sensor allows the robot to carefully position its end effector before using force control to manipulate an object.

\bibliography{references}
\bibliographystyle{IEEEtran}

\end{document}